\documentclass[11pt]{article}

\usepackage[final]{acl}

\usepackage{times}
\usepackage{latexsym}
\usepackage{amsmath}
\usepackage{amssymb}
\usepackage{graphicx}
\usepackage{booktabs}
\usepackage{multirow}
\usepackage{colortbl}
\usepackage{xcolor}
\usepackage{makecell}
\usepackage[most]{tcolorbox}
\definecolor{linecolor}{gray}{.90}
\definecolor{promptline}{HTML}{AFC6DD}
\definecolor{promptfill}{HTML}{FBFCFE}
\definecolor{prompttitle}{HTML}{DCEAF4}

\newtcolorbox{promptbox}[1]{
  breakable,
  enhanced,
  colback=promptfill,
  colframe=promptline,
  colbacktitle=prompttitle,
  coltitle=black,
  fonttitle=\bfseries,
  title=#1,
  boxrule=0.8pt,
  arc=2mm,
  left=2mm,
  right=2mm,
  top=1mm,
  bottom=1mm,
  fontupper=\small
}
\usepackage[T1]{fontenc}
\usepackage[utf8]{inputenc}
\usepackage{microtype}
\usepackage{inconsolata}

\usepackage{algorithm}
\usepackage{algorithmic}
\newcommand{\cc}[1]{\cellcolor{blue!10}#1}

\title{Mitigating Selection Bias in Large Language Models via Permutation-Aware GRPO}

\author{
  Jinquan Zheng, Jia Yuan, Jiacheng Yao, Chenyang Gu, Pujun Zheng, Guoxiu He\thanks{Corresponding author.} \\
  School of Economics and Management, East China Normal University \\
  \texttt{\{jqzheng, yuanjia, jcyao, cygu, pjzheng\}@stu.ecnu.edu.cn},\\
  \texttt{gxhe@fem.ecnu.edu.cn}
}

\begin{document}

\maketitle

\begin{abstract}
Large language models (LLMs) used for multiple-choice and pairwise evaluation tasks often exhibit selection bias due to non-semantic factors like option positions and label symbols. Existing inference-time debiasing is costly and may harm reasoning, while pointwise training ignores that the same question should yield consistent answers across permutations. To address this issue, we propose Permutation-Aware Group Relative Policy Optimization (PA-GRPO), which mitigates selection bias by enforcing permutation-consistent semantic reasoning. PA-GRPO constructs a permutation group for each instance by generating multiple candidate permutations, and optimizes the model using two complementary mechanisms: (1) cross-permutation advantage, which computes advantages relative to the mean reward over all permutations of the same instance, and (2) consistency-aware reward, which encourages the model to produce consistent decisions across different permutations. 
Experimental results demonstrate that PA-GRPO outperforms strong baselines across seven benchmarks, substantially reducing selection bias while maintaining high overall performance. The code is available on \href{https://github.com/ECNU-Text-Computing/PA-GRPO}{GitHub}.
\end{abstract}

\section{Introduction}
\label{sec:intro}

\begin{figure}[t!]
    \centering
    \includegraphics[trim=15.5cm 5.3cm 14cm 12cm, clip, width=1.05\linewidth]{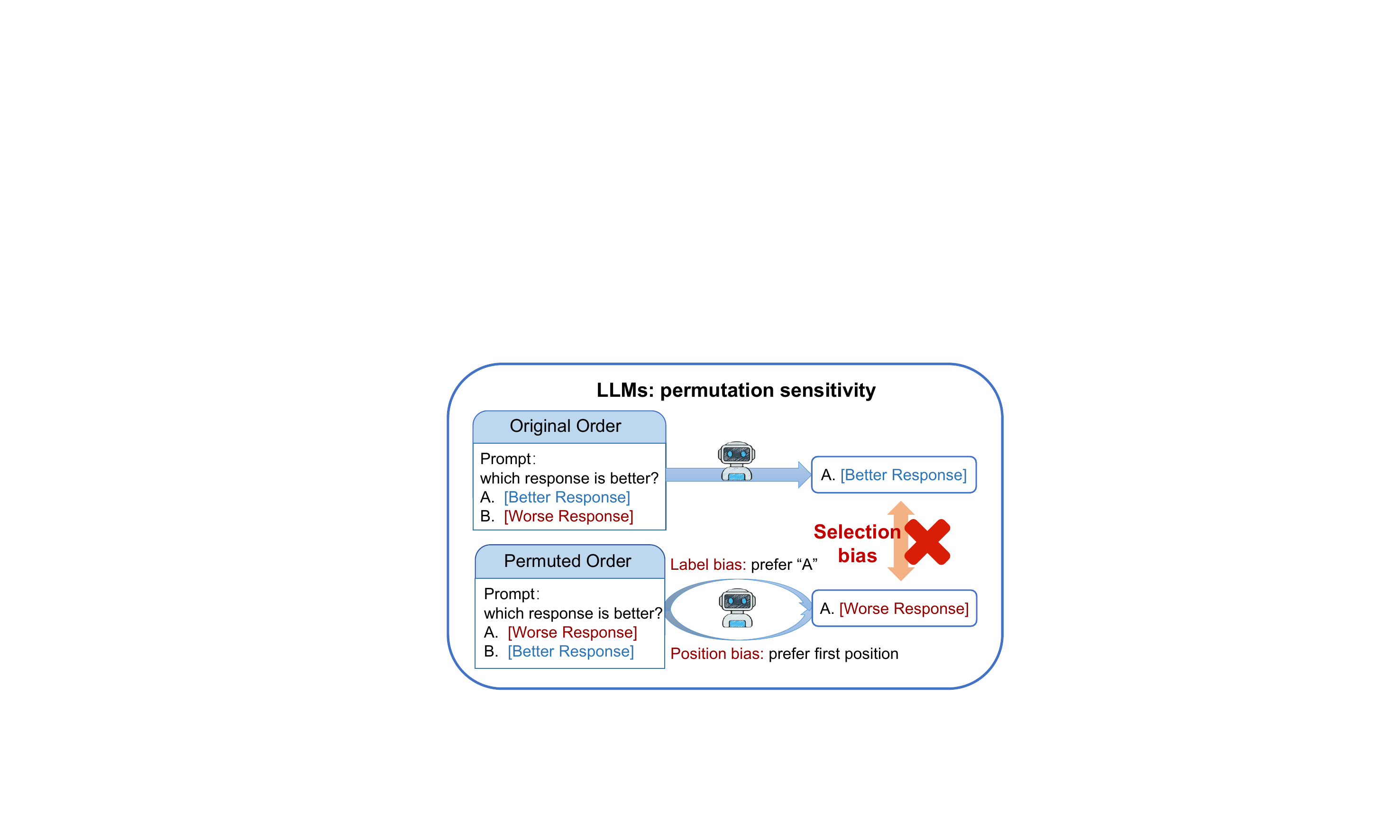}
    \caption{
    Example of selection bias in discrete-choice settings: swapping the order of answers changes only non-semantic factors.
    }
    \label{fig:motivation}
\end{figure}

Large language models (LLMs) have been widely applied to multiple-choice question (MCQ) reasoning~\cite{hendrycks2020measuring,achiam2023gpt} and LLM-as-a-Judge style evaluation tasks~\cite{zheng2023judging,chiang2024chatbot}. In these discrete-choice settings, option order and label symbols (\textit{e.g.}, A/B/C/D) are non-semantic factors that, in principle, should not influence the model's decisions. However, as illustrated in Figure~\ref{fig:motivation}, even when two candidate answers are swapped without altering their content, LLMs may still favor the first-listed option (\textit{i.e.}, \textbf{position bias}) or a particular label such as ``A'' (\textit{i.e.}, \textbf{label bias}), leading to inconsistent decisions and even preference reversals. Such systematic deviations caused by non-semantic perturbations are referred to as \textbf{selection bias}~\cite{zheng2023judging,zheng2023large}. Selection bias undermines the reliability and fairness of evaluation because model outputs can depend on the arrangement of options rather than their semantic content~\cite{shi2024judging,wei2024unveiling,wang2024large}. Moreover, these errors may be further amplified in downstream applications that rely on such outputs, such as automated data construction or alignment training~\cite{miao2024inform}.

Existing methods for mitigating selection bias can be broadly categorized into two types: inference-time debiasing and training-time debiasing.
Inference-time debiasing typically follows two routes. 
The first performs output-level calibration by estimating and removing priors post-hoc~\cite{zhao2021calibrate}, such as PriDe for MCQ and CalibraEval for LLM-as-a-Judge~\cite{zheng2023large,li2025calibraeval}. 
While often effective, such methods merely adjust surface-level probabilities without rectifying the model's intrinsic misconceptions, and typically incur high computational overhead due to repeated sampling or optimization.
The second route applies \textbf{internal interventions} to suppress bias sources, for instance by masking specific attention heads (UniBias) or pruning bias-related parameters (BNP)~\cite{zhou2024unibias,choi2025mitigating}. 
However, such invasive modifications complicate the inference pipeline and carry the risk of degrading general reasoning capabilities by inadvertently removing useful semantic features.
In contrast, training-time debiasing aims to internalize robustness. 
Yet, existing supervised strategies mostly rely on static augmentation or distillation~\cite{liusie2024teacher,xue2024strengthened}. 
These methods generally treat permutations as independent static samples, limiting the model to passive imitation of the data distribution rather than actively exploring the policy space to discover robust, permutation-invariant strategies.

Fundamentally, selection bias reflects a failure of robust reasoning in discrete-choice prompting: when only non-semantic factors (labels or positions) change, the model should preserve the same semantic decision.
Since this is an intrinsic reasoning flaw, a natural solution is to train the model to maintain consistency under such perturbations, rather than relying on post-hoc calibration at inference.
In this context, recent reinforcement learning (RL) methods such as Group Relative Policy Optimization (GRPO)~\cite{shao2024deepseekmath} are appealing, as they estimate relative advantages from multiple samples and can directly optimize decision quality.

However, standard GRPO and its variants still typically treat different prompts as independent training instances.
When multiple prompt variants are derived from the same semantic instance under different permutations, they are optimized separately without an explicit \textbf{cross-permutation consistency} constraint.
Consequently, the model may achieve high reward under a ``favorable'' permutation while performing poorly under others (\textit{e.g.}, reversed order), yet this inconsistency is not systematically penalized.
We term this failure mode permutation-blindness.

To address this, we propose \textbf{P}ermutation-\textbf{A}ware \textbf{G}roup \textbf{R}elative \textbf{P}olicy \textbf{O}ptimization (\textbf{PA-GRPO}), which explicitly targets \textbf{permutation invariance}: for all permuted variants of the same semantic instance, the model should make semantically consistent choices.
PA-GRPO organizes multiple permutations of the same instance into a \textbf{Permutation Group} and injects permutation awareness via two mechanisms:
(1) \textbf{Cross-Permutation Advantage}: we elevate the advantage baseline from the sample mean within a single prompt to the mean over all samples across the permutation group, so that positive advantages correspond to being globally better across permutations;
(2) \textbf{Consistency-Aware Reward}: we explicitly encourage consistent decisions across permutations in the reward, suppressing locally optimal strategies that exploit positional or label shortcuts.

We evaluate our approach on seven benchmarks spanning both MCQ and LLM-as-a-Judge tasks. During training, we construct representative permutation sets for each semantic instance, covering all label and position bindings, and validate on Llama-3.1-8B-Instruct, Qwen3-8B, and Qwen3-32B. Experimental results show that, compared with strong baselines such as PIF~\cite{xue2024strengthened} and CalibraEval~\cite{li2025calibraeval}, PA-GRPO achieves superior performance in both reducing selection bias and preserving stable performance.

Our contributions are threefold:

$\bullet$ We identify the permutation-blindness issue in RL alignment under selection-biased settings and propose permutation invariance as a training objective.
    
$\bullet$  We propose PA-GRPO, which incorporates consistency constraints into training through a cross-permutation advantage and a consistency-aware reward.
    
$\bullet$ Extensive experiments on three backbones and seven benchmarks verify that PA-GRPO effectively reduces selection bias.

\section{Related Work}
\label{sec:related_work}

\noindent \textbf{LLM-as-a-Judge and MCQ.}
With the rapid evolution of LLMs, efficient and reliable evaluation has become a central challenge.
A unifying perspective is that many widely used evaluation protocols can be formulated as discrete-choice decision making, where the model selects from a closed set of candidates.
One representative line is \textbf{LLM-as-a-Judge}~\cite{zheng2023judging}, which utilizes strong LLMs to perform pairwise comparisons of generated responses and has been widely adopted for alignment studies and leaderboard-style benchmarking~\cite{li2025generation}.
Another representative line is \textbf{Multiple Choice Question (MCQ)} evaluation~\cite{hendrycks2020measuring}, a standard format for assessing general reasoning by constraining the output space to predefined options.
Although their application scenarios differ, both settings ultimately reduce to assigning probabilities over a closed label set (\textit{e.g.}, A/B or A/B/C/D).
Crucially, this shared discrete-choice nature makes both settings inherently susceptible to non-semantic interference: even minor perturbations to the input form can trigger drastic fluctuations in decision outcomes, undermining the reliability of such discriminative tasks~\cite{pezeshkpour2024large,holtzman2021surface}.

\noindent \textbf{Selection Bias in LLMs.}
Models in discrete-choice tasks (\textit{e.g.}, LLM-as-a-Judge, MCQ) are sensitive to non-semantic factors, known as selection bias, including label bias and position bias. \textbf{Label bias}, preferring specific IDs, is addressed by approaches like SDC~\cite{wang2025beyond}, LOOC~\cite{reif2024beyond}, and ICC~\cite{jang2024rectifying} via content-free prompts or calibration, though often incurring extra inference overhead.
\textbf{Position bias}~\cite{lu2022fantastically}, preferring specific positions, is mitigated by mechanism adjustments: PINE~\cite{wang2024eliminating} and RoToR~\cite{yoon2025rotor} modify attention masks or positional IDs, while fine-tuning strategies~\cite{wang2025position} reduce sensitivity during training.

Mitigation strategies for selection bias fall into two categories.
\textbf{Inference-time debiasing} includes calibration methods like PriDe~\cite{zheng2023large} and CalibraEval~\cite{li2025calibraeval}, which remove priors post-hoc without fixing the underlying mechanism.
Alternatively, internal interventions such as UniBias~\cite{zhou2024unibias} and BNP~\cite{choi2025mitigating} prune bias sources but risk unintended side effects.
In contrast, \textbf{training-time debiasing} aims to internalize robustness.
However, existing supervised strategies~\cite{liusie2024teacher,xue2024strengthened} typically treat permutations as static samples, limiting models to passive imitation rather than actively exploring the policy space to discover robust, permutation-invariant strategies.

\noindent \textbf{Reinforcement Learning for Alignment.}
Reinforcement Learning (RL)~\cite{christiano2017deep} has become a core technology for alignment, evolving from RLHF (PPO)~\cite{ouyang2022training} to DPO~\cite{rafailov2023direct}, and the efficient GRPO~\cite{shao2024deepseekmath} with its variants~\cite{yu2025dapo,zheng2025group}.
While these methods significantly improve generation quality, they fundamentally rely on a sample-independence assumption.
Even group-based methods like GRPO typically treat permuted variants of the same semantic input as unrelated, independent events.
This leads to permutation-blindness: the optimization focuses on maximizing reward for a specific order in isolation, ignoring consistency across permutations.
Consequently, existing RL frameworks fail to enforce global consistency across input variations, a critical limitation that our work seeks to address.

\begin{figure*}[t!]
    \centering
    \includegraphics[trim=1.5cm 7cm 6cm 1.5cm, clip, width=\linewidth]{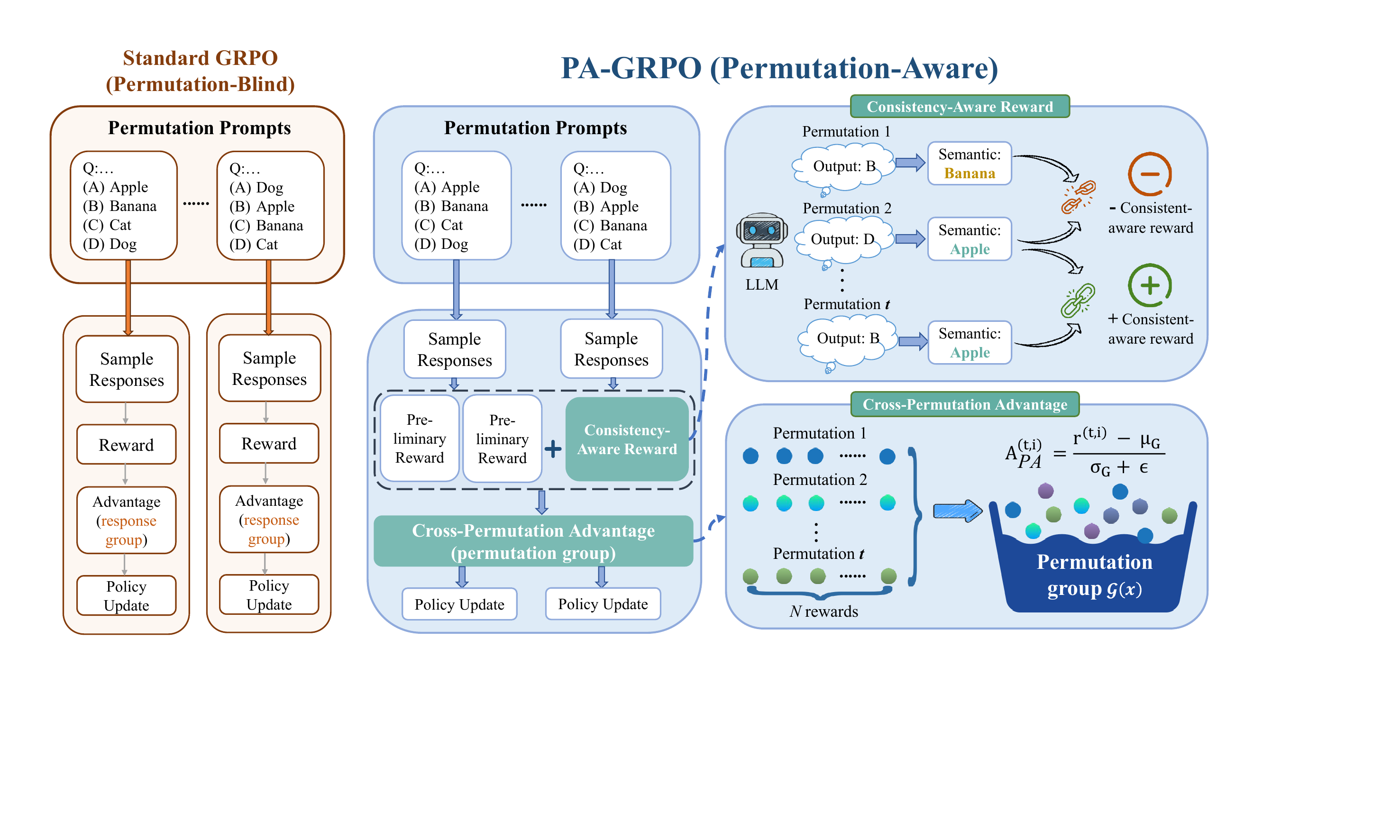}
    \caption{
        Comparison between \textbf{Standard GRPO} (left) and the proposed \textbf{PA-GRPO} (right).
        Standard GRPO treats permuted prompts as independent samples, suffering from permutation blindness where inconsistency goes unpunished.
        In contrast, PA-GRPO organizes samples into Permutation Groups. It introduces (1) a Cross-Permutation Advantage (using the permutation group mean as a baseline) and (2) a Consistency-Aware Reward to explicitly enforce semantic invariance across different permutations of the same instance.
        }
    \label{fig:method}
\end{figure*}

\section{Methodology}
\label{sec:method}

We propose \textbf{PA-GRPO} (Permutation-Aware Group Relative Policy Optimization) to mitigate selection bias induced by non-semantic factors in discrete-choice reasoning tasks. 
As illustrated in \textbf{Figure \ref{fig:method}}, unlike standard GRPO which treats each prompt variation independently, PA-GRPO explicitly models the dependency among permutation variants.
Our core objective is to jointly optimize for two goals: (1) maximizing semantic correctness to align with the ground truth, and (2) enforcing permutation invariance to maintain consistent decisions regardless of option order.

\subsection{Problem Formulation}
\label{sec:formulation}
A base instance $x$ consists of a question $q$ and a candidate set $\mathcal{C}$.
We unify MCQ ($|\mathcal{C}|=4$) and Pairwise Judge ($|\mathcal{C}|=2$) as conditional generation tasks. Given a prompt $p$, a policy $\pi_\theta(y\mid p)$ generates a response $y$. We parse the surface label $\ell(y)\in\mathcal{L}$ (\textit{e.g.}, A/B/C/D) from the output.
Ideally, a robust model should satisfy permutation invariance: for any non-semantic permutation $\tau$, the mapped semantic choice should remain invariant. Selection bias is defined as the violation of this property, manifested as systematic preference flips triggered by label or position changes.

\subsection{Permutation Group Construction and Sampling}
\label{sec:perm_group}
To enforce cross-permutation consistency, we organize training data into \textbf{Permutation Groups}.
For each base instance $x$, we apply a set of permutation mappings $\Pi$ to generate $P$ distinct prompt variants:
\begin{equation}
\mathcal{G}(x)=\{p^{(t)}=\tau_t(x)\}_{t=1}^{P},\quad \tau_t\in\Pi.
\end{equation}
For each permuted prompt $p^{(t)}$, we further sample a \textbf{response group} of $N$ outputs $\{y^{(t,i)}\}_{i=1}^{N}$ from the current policy $\pi_\theta$ (as in GRPO).
Together, this yields a two-level structure: sampling diversity within each permutation and robustness across permutations.
In the following sections, we compute rewards and advantage baselines over all $P\times N$ samples, enabling explicit cross-permutation learning and reducing reliance on any single ``favorable'' permutation.
Specifically, we design the permutation set $\Pi$ for different tasks as follows:

\noindent \textbf{MCQ ($P=5$, Structured Subset).}
The full permutation space ($4! = 24$) is costly in computation.
We employ a structured subset strategy consisting of four cyclic shifts and one reverse order:
\begin{equation}
\Pi_{\text{MCQ}}=\{\text{ABCD},\text{BCDA},\text{CDAB},\text{DABC},\text{DCBA}\}.
\end{equation} 
The four cyclic shifts ensure strictly uniform coverage of each option at every position. We additionally include the reverse order (\texttt{DCBA}) to disrupt relative adjacency patterns (\textit{e.g.}, A is always before B in cyclic shifts), thereby enhancing robustness against complex order-dependent biases.

\noindent \textbf{\textsc{Judge} ($P=2$, Full Set).}
For pairwise comparison, the space is small ($2!=2$). We employ the full symmetric set:
\begin{equation}
\Pi_{\text{Judge}}=\{\text{AB},\text{BA}\}.
\end{equation}

\subsection{Reward Design: Content Alignment and Consistency Awareness}
\label{sec:reward}

Standard RLHF~\cite{ouyang2022training} typically relies on a scalar reward signal reflecting response correctness. 
However, in the context of debiasing, a correct response under a specific permutation is insufficient; the model must demonstrate robust correctness across all variations.
Therefore, we design a composite reward mechanism that explicitly decouples the evaluation into two dimensions:
(1) \textbf{Preliminary Reward} ($r_{\text{pre}}$), which assesses whether the current response matches the ground truth;
(2) \textbf{Consistency Reward} ($r_{\text{con}}$), which penalizes semantic divergence within the permutation group.
To implement this, we first establish a mapping from surface labels to semantic identities.

\subsubsection{Content Alignment: From Labels to Semantic Choices}
Consistency must be defined over the selected content, not the surface letters.
For each permutation mapping $\tau_t$, we define a mapping function $m_{\tau_t}$ that maps a surface label to the index of the original semantic candidate:
\begin{equation}
m_{\tau_t}(\ell)\in\{1,\dots,|\mathcal{C}|\}.
\end{equation}
For the $i$-th sampled response under permutation $t$, the semantic choice is:
\begin{equation}
z^{(t,i)} = m_{\tau_t}(\ell(y^{(t,i)})).
\end{equation}

\subsubsection{Preliminary Reward }
For each sampled response $y^{(t,i)}$, the total reward is:
\begin{equation}
r^{(t,i)} = r_{\text{pre}}^{(t,i)} + \lambda\, r_{\text{con}}^{(t,i)}.
\label{eq:total_reward}
\end{equation}
The preliminary reward aggregates accuracy, length, and format constraints:
\begin{equation}
r_{\text{pre}}^{(t,i)} = r_{\text{acc}}^{(t,i)} + r_{\text{len}}^{(t,i)} + r_{\text{fmt}}^{(t,i)}.
\end{equation}
In our implementation, we use: correctness reward $r_{\text{acc}}\in\{+1,-1\}$ based on whether $z^{(t,i)}$ matches the ground-truth semantic answer, along with minor regularization terms for length ($r_{\text{len}}=\pm 0.1$) and formatting ($r_{\text{fmt}}=\pm 0.3$).

\subsubsection{Consistency-Aware Reward}
We introduce a consistency reward $r_{\text{con}}$ to encourage semantically consistent decisions within each permutation group $\mathcal{G}(x)$.

\noindent \textbf{\textsc{Judge} ($P=2$): index-aligned pairwise consistency.}
We pair the $i$-th samples across the two permutations:
\begin{equation}
r_{\text{con}}^{(1,i)} = r_{\text{con}}^{(2,i)}=
\begin{cases}
+1, & z^{(1,i)} = z^{(2,i)}\\
-1, & z^{(1,i)} \neq z^{(2,i)}.
\end{cases}
\label{eq:judge_cons}
\end{equation}

\noindent \textbf{MCQ ($P=5$): unique-mode agreement.}
Let $n_k$ be the count of semantic choice $k$ across the entire group, and $\mathcal{M}$ be the set of modes:
\begin{equation}
\begin{aligned}
n_k &= \sum_{t=1}^{P}\sum_{i=1}^{N}\mathbf{1}\big[z^{(t,i)}=k\big], \\
\mathcal{M} &= \arg\max_{k} n_k.
\end{aligned}
\end{equation}
If $|\mathcal{M}|=1$, denote the unique mode by $z^\star$.
We explicitly penalize fragmentation (including ties):
\begin{equation}
r_{\text{con}}^{(t,i)}=
\begin{cases}
+1, & |\mathcal{M}|=1\ \text{and}\ z^{(t,i)}=z^\star\\
-1, & \text{otherwise}.
\end{cases}
\label{eq:mcq_cons}
\end{equation}

\subsection{Cross-Permutation Advantage Estimation}
\label{sec:adv}

Prompts in a permutation group $\mathcal{G}(x)$ are semantically identical but differ in non-semantic surface factors.
If advantages are computed independently per prompt variant (as in standard GRPO), the policy can be reinforced by a ``favorable'' ordering even when it fails under other permutations.

PA-GRPO therefore lifts advantage estimation to the \textbf{permutation-group level}.
We treat the $P\times N$ samples from $\mathcal{G}(x)$ as a single comparison set and compute the permutation-group mean and standard deviation:
\begin{equation}
\begin{aligned}
\mu_{\mathcal{G}} &= \frac{1}{PN}\sum_{t=1}^{P}\sum_{i=1}^{N} r^{(t,i)}, \\
\sigma_{\mathcal{G}} &= \operatorname{Std}\big(\{r^{(t,i)}\}_{t,i}\big),
\end{aligned}
\end{equation}
where both statistics are computed over all permutations and samples within $\mathcal{G}(x)$.
The cross-permutation advantage is:
\begin{equation}
A_{\text{PA}}^{(t,i)}=
\begin{cases}
0, & \sigma_{\mathcal{G}}<\delta\\
\dfrac{r^{(t,i)}-\mu_{\mathcal{G}}}{\sigma_{\mathcal{G}}+\epsilon}, & \text{otherwise}.
\end{cases}
\label{eq:adv}
\end{equation}

This yields (i) \textbf{instance-wise competition} across permutations via the permutation-group baseline $\mu_{\mathcal{G}}$: a sample receives a positive advantage only if its reward exceeds the permutation-group mean, and (ii) \textbf{scale-invariant stability} via normalization by $\sigma_{\mathcal{G}}$. When $\sigma_{\mathcal{G}}<\delta$, rewards are nearly indistinguishable within the group, so we set $A_{\text{PA}}=0$ to avoid amplifying noise.

\subsection{Optimization Objective}
\label{sec:objective}
We adopt a clipped objective with KL regularization to stabilize training. 
Let the importance ratio be defined as:
\begin{equation}
    \rho^{(t,i)}(\theta) = \frac{\pi_\theta(y^{(t,i)} \mid p^{(t)})}{\pi_{\text{old}}(y^{(t,i)} \mid p^{(t)})}
    \label{eq:rho}
\end{equation}
The clipped surrogate objective is:
\begin{multline}
\mathcal{L}_{\text{clip}}(\theta)
= \mathbb{E}\Big[
\min\Big(
\rho^{(t,i)}A_{\text{PA}}^{(t,i)}, \\
\operatorname{clip}\!\big(\rho^{(t,i)},\, 1-\eta,\, 1+\eta\big)
A_{\text{PA}}^{(t,i)}
\Big)
\Big].
\label{eq:lclip}
\end{multline}

The final objective maximizes the surrogate reward while penalizing deviation from the reference policy:
\begin{multline}
\max_\theta\ \Big( \mathcal{L}_{\text{clip}}(\theta) \ - \\
\beta\, \mathbb{E}_{p^{(t)}}\Big[\operatorname{KL}\!\big(\pi_\theta(\cdot| p^{(t)}) \parallel \pi_{\text{ref}}(\cdot| p^{(t)})\big)\Big] \Big).
\label{eq:full_obj}
\end{multline}
The complete training procedure is provided in \textbf{Algorithm~\ref{alg:pa_grpo}} in Appendix~\ref{app:algorithm}.

\section{Experimental Setup}
\label{sec:setup}
In this section, we present a comprehensive evaluation of PA-GRPO to answer the following research questions (RQs): 

\textbf{RQ1:} Can PA-GRPO outperform existing inference-time calibration and training-time debiasing methods?

\textbf{RQ2:} What are the individual contributions of the Cross-Permutation Advantage ($A_{\text{PA}}$) and the Consistency-Aware Reward ($r_{\text{con}}$)?

\textbf{RQ3:} How do the consistency reward coefficient $\lambda$ and permutation group size $P$ affect model performance, permutation invariance, and training efficiency?

\textbf{RQ4:} Can PA-GRPO effectively align reasoning consistency under Chain-of-Thought prompting?

\textbf{RQ5:} How does PA-GRPO's robustness differ between isolated label-only and order-only perturbations and the naturally coupled permutations used in training, and what does this reveal about the remaining sources of bias?

\subsection{Datasets and Evaluation Protocol}
\label{sec:datasets}

We evaluated PA-GRPO on seven benchmarks spanning two categories: (1) \textbf{\textsc{LLM-as-a-Judge}}: MT-Bench~\cite{zheng2023judging}, JudgeBench~\cite{tan2024judgebench}, PreferenceBench~\cite{kim2024prometheus}, and RewardBench~\cite{lambert2025rewardbench}; and (2) \textbf{MCQ}: ARC-Challenge~\cite{clark2018think}, TinyMMLU~\cite{polo2024tinybenchmarks}, and GPQA~\cite{rein2024gpqa}. Brief descriptions of each benchmark and their task formats are provided in Appendix~\ref{app:benchmarks}.
For training, we utilized the Chatbot Arena dataset~\cite{zheng2023judging} for pairwise tasks and the MMLU training set~\cite{hendrycks2020measuring} for reasoning tasks. 
Details of data construction are provided in Appendix~\ref{app:data_construction}.

To rigorously quantify permutation invariance during inference, we employed \textbf{Full Permutation Expansion} rather than random sampling. 
Specifically, we evaluated the complete permutation space for every test instance, covering all $2!=2$ possible orders for LLM-as-a-Judge tasks and all $4!=24$ possible arrangements for MCQ tasks.

\subsection{Metrics}
\label{sec:metrics}

We report three core metrics to measure performance and robustness:

\noindent\textbf{Accuracy (Acc).} 
The standard performance metric, calculated as the average correctness over all samples in the expanded permutation groups.

\noindent\textbf{Consistency (Con).} 
It measures decision stability across permutations. 
For LLM-as-a-Judge, it is defined as the proportion of instance pairs where the model makes the identical semantic choice under both orders. 
For MCQ, we define it as the majority consensus ratio $n_{\text{mode}}/24$, where $n_{\text{mode}}$ is the count of the most frequent semantic choice within the permutation group.

\noindent\textbf{Consistent Accuracy (CA).} 
It measures the ability to be both consistent and correct. 
For LLM-as-a-Judge, we apply a strict criterion: an instance scores 1 if and only if the model answers correctly under both permutations. 
For MCQ, we adopt Majority Vote Correctness~\cite{wang2022self}: an instance scores 1 if the mode of the permutation group matches the ground truth. This reflects whether the consensus converges to the truth after filtering out permutation noise.

\newcommand{\good}[2]{\makecell[c]{\textbf{#1} \\[-0.35em] \textcolor{teal}{\scriptsize{(+#2)}}}} 
\newcommand{\bad}[2]{\makecell[c]{#1 \\[-0.35em] \textcolor{red}{\scriptsize{(#2)}}}}           

\begin{table*}[t]
\centering
\resizebox{\textwidth}{!}{
\setlength{\aboverulesep}{0pt}
\setlength{\belowrulesep}{0pt}
\setlength{\tabcolsep}{5.5pt} 
\renewcommand{\arraystretch}{1.2}
\begin{tabular}{c|l|ccc|ccc|ccc|ccc|ccc|ccc|ccc}
\toprule
& & \multicolumn{12}{c}{\textbf{LLM-as-a-Judge}} & \multicolumn{9}{c}{\textbf{Multiple Choice Questions}} \\
\cmidrule(lr){3-14} \cmidrule(lr){15-23}
\multirow{2}{*}{\textbf{Model}} & \multirow{2}{*}{\textbf{Method}} & \multicolumn{3}{c|}{\textbf{MT-Bench}} & \multicolumn{3}{c|}{\textbf{JudgeBench}} & \multicolumn{3}{c|}{\textbf{PreferenceBench}} & \multicolumn{3}{c|}{\textbf{RewardBench}} & \multicolumn{3}{c|}{\textbf{GPQA}} & \multicolumn{3}{c|}{\textbf{ARC-Challenge}} & \multicolumn{3}{c}{\textbf{TinyMMLU}} \\
& & Acc & Con & CA & Acc & Con & CA & Acc & Con & CA & Acc & Con & CA & Acc & Con & CA & Acc & Con & CA & Acc & Con & CA \\
\midrule
\multirow{7}{*}{\rotatebox[origin=c]{90}{\textbf{Llama-3.1-8B}}} 
& Base & 59.6 & 25.2 & 22.2 & 35.0 & 34.8 & 6.1 & 60.8 & 22.6 & 22.1 & 60.5 & 31.5 & 26.2 & 27.3 & 40.0 & 27.2 & 73.0 & 82.5 & 78.9 & 54.2 & 68.9 & 57.0 \\
& GRPO & 75.7 & 80.6 & 65.4 & 48.2 & 56.1 & 28.2 & 82.2 & 85.1 & 76.3 & 70.9 & 76.9 & 61.5 & 29.5 & 55.6 & 29.7 & 89.0 & 92.1 & 91.5 & 67.3 & 77.9 & 73.0 \\
& UniBias & 68.5 & 59.4 & 48.2 & 50.2 & 23.0 & 10.9 & 79.7 & 65.6 & 62.5 & 68.4 & 64.4 & 50.6 & \textbf{32.3} & 56.1 & 31.8 & 80.4 & 90.0 & 83.6 & 61.1 & 76.5 & 65.0 \\
& PriDe & 65.6 & 46.2 & 38.9 & 49.1 & 16.2 & 7.2 & 81.6 & 69.1 & 66.2 & 65.1 & 52.8 & 41.5 & 31.0 & 49.1 & 27.4 & 79.1 & 83.0 & 79.0 & 59.7 & 69.5 & 58.0 \\
& CalibraEval & 62.3 & 42.1 & 33.4 & 49.3 & 15.7 & 7.1 & 70.4 & 44.4 & 42.7 & 60.7 & 34.4 & 27.8 & 28.3 & 48.2 & 26.5 & 74.4 & 71.0 & 68.3 & 55.6 & 72.1 & 62.0 \\
& PIF & 76.1 & 84.6 & 70.4 & 53.3 & \textbf{59.2} & 30.4 & 54.3 & 37.1 & 25.7 & \textbf{73.7} & 76.7 & 62.0 & 29.1 & 49.7 & 29.2 & 73.3 & 83.0 & 78.8 & 56.9 & 69.1 & 57.0 \\
\cellcolor{white} & \cc PA-GRPO & \rule{0pt}{3.5ex}\cc\good{77.6}{1.5} & \cc\good{88.0}{3.4} & \cc\good{71.7}{1.3} & \cc\good{57.1}{3.8} & \cc\bad{58.3}{-0.9} & \cc\good{32.4}{2.0} & \cc\good{86.2}{4.0} & \cc\good{87.2}{2.1} & \cc\good{79.8}{3.5} & \cc\bad{71.0}{-2.7} & \cc\good{82.7}{5.8} & \cc\good{62.3}{0.3} & \cc\bad{32.1}{-0.2} & \cc\good{57.2}{1.1} & \cc\good{33.2}{1.4} & \cc\good{89.3}{0.3} & \cc\good{92.9}{0.8} & \cc\good{92.6}{1.1} & \cc\good{67.9}{0.6} & \cc\good{81.9}{4.0} & \cc\good{75.0}{2.0} \\
\midrule
\multirow{7}{*}{\rotatebox[origin=c]{90}{\textbf{Qwen3-8B}}} 
& Base & 71.2 & 56.3 & 49.4 & 43.9 & 45.5 & 16.5 & 82.3 & 70.2 & 67.4 & 79.3 & 73.8 & 66.2 & 32.5 & 48.2 & 30.4 & 90.5 & 94.7 & 92.4 & 67.4 & 79.6 & 72.0 \\
& GRPO & 78.1 & 82.5 & 69.3 & 50.4 & 62.6 & 34.8 & 82.9 & 77.3 & 76.2 & \textbf{86.8} & 86.7 & 80.2 & 40.4 & 63.2 & 43.8 & 94.6 & 97.2 & 95.9 & 56.6 & 83.7 & 76.0 \\
& UniBias & 67.0 & 52.7 & 43.4 & 52.2 & 26.1 & 14.9 & 75.8 & 55.4 & 53.5 & 66.6 & 45.9 & 39.5 & 34.6 & 54.3 & 34.2 & 89.9 & 94.1 & 91.8 & 71.3 & 81.0 & 74.0 \\
& PriDe & 72.6 & 66.7 & 52.1 & 51.2 & 48.8 & 29.8 & 81.5 & 72.6 & 69.7 & 80.9 & 79.1 & 68.7 & 34.4 & 56.9 & 38.5 & 91.2 & 94.9 & 92.6 & 68.9 & 79.9 & 78.0 \\
& CalibraEval & 72.4 & 62.1 & 50.9 & 49.7 & 56.4 & 31.3 & 76.5 & 72.8 & 70.1 & 80.4 & 71.2 & 69.5 & 33.5 & 49.8 & 32.9 & 90.1 & 91.0 & 89.8 & 63.5 & 75.5 & 68.0 \\
& PIF & 75.1 & 83.2 & 68.9 & 54.3 & 59.6 & 37.4 & 77.7 & 64.2 & 58.9 & 83.9 & 80.1 & 75.8 & 32.7 & 68.7 & 46.6 & 92.2 & 95.4 & 93.5 & \textbf{75.1} & 81.4 & 75.0 \\
\cellcolor{white} & \cc PA-GRPO & \rule{0pt}{3.5ex}\cc\good{78.8}{0.7} & \cc\good{86.4}{3.2} & \cc\good{72.0}{2.7} & \cc\good{60.1}{5.8} & \cc\good{70.0}{7.4} & \cc\good{45.3}{7.9} & \cc\good{88.9}{6.0} & \cc\good{89.6}{\tiny 12.3} & \cc\good{83.7}{7.5} & \cc\bad{84.4}{-2.4} & \cc\good{88.9}{2.2} & \cc\good{80.6}{0.4} & \cc\good{42.3}{1.9} & \cc\good{72.4}{3.7} & \cc\good{56.7}{\tiny 10.1} & \cc\good{95.0}{0.4} & \cc\good{97.5}{0.3} & \cc\good{96.2}{0.3} & \cc\bad{73.2}{-1.9} & \cc\good{88.6}{4.9} & \cc\good{80.0}{2.0} \\
\midrule
\multirow{7}{*}{\rotatebox[origin=c]{90}{\textbf{Qwen3-32B}}} 
& Base & 78.5 & 83.4 & 70.3 & 55.4 & 62.1 & 29.7 & 84.9 & 85.1 & 78.7 & 85.4 & 84.3 & 77.5 & 40.4 & 59.9 & 46.7 & 94.5 & 97.2 & 95.7 & 76.5 & 86.0 & 81.0 \\
& GRPO & 81.0 & 90.6 & 73.7 & 61.4 & 74.2 & \textbf{45.1} & 89.1 & 90.3 & 83.3 & 88.2 & 91.4 & 83.9 & 53.2 & 74.3 & 59.9 & 95.3 & 98.1 & 97.0 & 85.2 & 91.4 & 86.0 \\
& UniBias & 76.5 & 85.6 & 71.1 & 56.2 & 64.0 & 32.5 & 87.2 & 91.1 & 84.1 & 85.7 & 85.9 & 78.0 & 46.9 & 68.9 & 54.5 & 94.5 & 95.8 & 95.1 & 81.0 & 89.3 & 80.0 \\
& PriDe & 77.2 & 86.1 & 72.1 & 56.8 & 63.5 & 33.1 & 85.1 & 86.7 & 81.2 & 86.9 & 88.2 & 80.3 & 49.7 & 73.1 & 58.5 & 95.5 & 97.4 & 95.8 & 79.9 & 84.1 & 84.0 \\
& CalibraEval & 78.8 & 79.1 & 71.2 & 52.9 & 61.0 & 28.9 & 83.7 & 87.2 & 79.9 & 85.5 & 86.7 & 79.6 & 48.0 & 72.2 & 55.2 & 93.9 & 96.8 & 96.0 & 80.3 & 83.3 & 85.0 \\
& PIF & 79.4 & 86.5 & 72.8 & \textbf{62.2} & 68.4 & 36.9 & 89.3 & 90.6 & 85.5 & \textbf{90.8} & 90.9 & 84.7 & 52.1 & 72.8 & 60.1 & \textbf{96.1} & 97.6 & 96.8 & 84.3 & \textbf{92.0} & 88.0 \\
\rowcolor{blue!10}
\cellcolor{white} & PA-GRPO & \rule{0pt}{3.5ex}\good{81.4}{0.4} & \good{91.6}{1.0} & \good{74.8}{1.1} & \bad{59.4}{-2.8} & \good{75.2}{1.0} & \bad{43.4}{-1.7} & \good{90.2}{0.9} & \good{92.1}{1.0} & \good{86.2}{0.7} & \bad{90.5}{-0.3} & \good{92.1}{0.7} & \good{86.6}{1.9} & \good{54.1}{0.9} & \good{75.1}{0.8} & \good{61.2}{1.1} & \bad{96.0}{-0.1} & \good{98.3}{0.2} & \good{97.2}{0.2} & \good{86.8}{1.6} & \bad{91.8}{-0.2} & \good{89.0}{1.0} \\
\bottomrule
\end{tabular}
}
\caption{Main results on Llama-3.1-8B-Instruct, Qwen3-8B, and Qwen3-32B across 7 benchmarks. All scores are rounded to one decimal place. PA-GRPO results show the absolute score (top) and the gain/drop compared to the best baseline (bottom). \textbf{Bold} indicates the best result in each column.}
\label{tab:main_results}
\end{table*}

\subsection{Baselines}
\label{sec:baselines}
We compare PA-GRPO with five strong baselines covering both inference-time debiasing and training-time alignment.
Inference-time methods include PriDe~\cite{zheng2023large}, which estimates and removes a global prior preference over option-ID tokens; CalibraEval~\cite{li2025calibraeval}, which learns a label-free calibration mapping to enforce swap-invariant judging; and UniBias~\cite{zhou2024unibias}, which mitigates bias by masking identified biased attention heads and FFN vectors.
Training-time baselines include PIF~\cite{xue2024strengthened}, an SFT algorithm that strengthens multiple-choice symbol binding via point-wise feedback on constructed negative instances, and standard GRPO~\cite{shao2024deepseekmath}.
Due to space constraints, detailed mechanisms and our implementation settings are provided in Appendix~\ref{app:baselines}.

\subsection{Implementation Details}
\label{sec:implementation}
We used Llama-3.1-8B-Instruct~\cite{dubey2024llama}, Qwen3-8B, and Qwen3-32B~\cite{yang2025qwen3} as policy models.
All models were trained within the \textsc{verl}~\cite{sheng2025hybridflow} RL framework, and we employed LoRA~\cite{hu2022lora} for fine-tuning.
Detailed training configurations and hyperparameters are provided in Appendix~\ref{app:hyperparams}.

\section{Results and Analysis}
\label{sec:results}

\begin{table*}[t]
\centering
\resizebox{\textwidth}{!}{
\setlength{\aboverulesep}{0pt}
\setlength{\belowrulesep}{0pt}
\setlength{\tabcolsep}{4pt}  
\renewcommand{\arraystretch}{1.2}
\begin{tabular}{l|ccc|ccc|ccc|ccc|ccc|ccc|ccc}
\toprule
& \multicolumn{12}{c}{\textbf{LLM-as-a-Judge}} & \multicolumn{9}{c}{\textbf{Multiple Choice Questions}} \\
\cmidrule(lr){2-13} \cmidrule(lr){14-22}
\multirow{2}{*}{\textbf{Method}} & \multicolumn{3}{c|}{\textbf{MT-Bench}} & \multicolumn{3}{c|}{\textbf{JudgeBench}} & \multicolumn{3}{c|}{\textbf{PreferenceBench}} & \multicolumn{3}{c|}{\textbf{RewardBench}} & \multicolumn{3}{c|}{\textbf{GPQA}} & \multicolumn{3}{c|}{\textbf{ARC-Challenge}} & \multicolumn{3}{c}{\textbf{TinyMMLU}} \\
& Acc & Con & CA & Acc & Con & CA & Acc & Con & CA & Acc & Con & CA & Acc & Con & CA & Acc & Con & CA & Acc & Con & CA \\
\midrule
Base & 59.6 & 25.2 & 22.2 & 35.0 & 34.8 & 6.1 & 60.8 & 22.6 & 22.1 & 60.5 & 31.5 & 26.2 & 27.3 & 40.0 & 27.2 & 73.0 & 82.5 & 78.9 & 54.2 & 68.9 & 57.0 \\
GRPO & 75.7 & 80.6 & 65.4 & 48.2 & 56.1 & 28.2 & 82.2 & 85.1 & 76.3 & 70.9 & 76.9 & 61.5 & 29.5 & 55.6 & 29.7 & 89.0 & 92.1 & 91.5 & 67.3 & 77.9 & 73.0 \\
+ $r_{\text{con}}$ & 75.2 & 83.1 & 66.5 & 48.0 & 56.5 & 29.2 & 82.6 & 85.9 & 76.9 & 70.6 & 79.0 & 61.9 & 29.9 & 56.4 & 31.0 & 87.8 & 92.3 & 92.0 & 67.1 & 78.8 & 73.0 \\
+ $A_{\text{PA}}$ & 76.6 & 84.2 & 67.8 & 49.3 & 55.4 & 30.5 & 83.4 & 86.4 & 77.8 & 69.8 & 78.7 & 61.1 & 30.7 & 57.1 & 32.2 & \textbf{89.7} & 92.3 & 91.9 & 67.5 & 79.4 & 74.0 \\
\rowcolor{blue!10}
PA-GRPO & \textbf{77.6} & \textbf{88.0} & \textbf{71.7} & \textbf{57.1} & \textbf{58.3} & \textbf{32.4} & \textbf{86.2} & \textbf{87.2} & \textbf{79.8} & \textbf{71.0} & \textbf{82.7} & \textbf{62.3} & \textbf{32.1} & \textbf{57.2} & \textbf{33.2} & 89.3 & \textbf{92.9} & \textbf{92.6} & \textbf{67.9} & \textbf{81.9} & \textbf{75.0} \\
\bottomrule
\end{tabular}
}
\caption{Ablation study on Llama-3.1-8B-Instruct across all 7 benchmarks. We group datasets into LLM-as-a-Judge tasks and MCQ tasks. The best results are \textbf{bolded}. Adding both components ($A_{\text{PA}}$ and $r_{\text{con}}$) consistently yields the best robustness.}
\label{tab:ablation}
\end{table*}

\begin{table*}[t]
\centering
\resizebox{\textwidth}{!}{
\setlength{\tabcolsep}{3.5pt}  
\renewcommand{\arraystretch}{1.3}
\begin{tabular}{l|ccc|ccc|ccc|ccc|ccc|ccc|ccc}
\toprule
& \multicolumn{12}{c}{\textbf{LLM-as-a-Judge}} & \multicolumn{9}{c}{\textbf{Multiple Choice Questions}} \\
\cmidrule(lr){2-13} \cmidrule(lr){14-22}
\multirow{2}{*}{\textbf{Setting}} & \multicolumn{3}{c|}{\textbf{MT-Bench}} & \multicolumn{3}{c|}{\textbf{JudgeBench}} & \multicolumn{3}{c|}{\textbf{PreferenceBench}} & \multicolumn{3}{c|}{\textbf{RewardBench}} & \multicolumn{3}{c|}{\textbf{GPQA}} & \multicolumn{3}{c|}{\textbf{ARC-Challenge}} & \multicolumn{3}{c}{\textbf{TinyMMLU}} \\
& Acc & Con & CA & Acc & Con & CA & Acc & Con & CA & Acc & Con & CA & Acc & Con & CA & Acc & Con & CA & Acc & Con & CA \\
\midrule
Base & 59.6 & 25.2 & 22.2 & 35.0 & 34.8 & 6.1 & 60.8 & 22.6 & 22.1 & 60.5 & 31.5 & 26.2 & 27.3 & 40.0 & 27.2 & 73.0 & 82.5 & 78.9 & 54.2 & 68.9 & 57.0 \\
$\lambda=0.5$ & \textbf{78.0} & 86.4 & 69.8 & 56.5 & 55.4 & 31.9 & 85.3 & 86.8 & 78.9 & 70.7 & 80.9 & 60.5 & \textbf{32.1} & 56.6 & 32.0 & \textbf{89.4} & 92.0 & 92.4 & 67.1 & 78.2 & 74.0 \\
\rowcolor{blue!10}
$\lambda=1.0$ (Default) & 77.6 & 88.0 & \textbf{71.7} & \textbf{57.1} & \textbf{58.3} & \textbf{32.4} & \textbf{86.2} & 87.2 & \textbf{79.8} & \textbf{71.0} & \textbf{82.7} & \textbf{62.3} & \textbf{32.1} & 57.2 & 33.2 & 89.3 & \textbf{92.9} & \textbf{92.6} & \textbf{67.9} & \textbf{81.9} & \textbf{75.0} \\
$\lambda=2.0$ & 76.5 & \textbf{88.6} & 70.5 & 56.8 & 55.9 & 32.0 & 85.0 & \textbf{87.5} & 79.1 & 70.3 & 82.5 & 62.1 & 31.4 & \textbf{58.0} & \textbf{33.6} & 88.7 & 91.2 & 91.1 & 66.4 & 80.1 & 73.0 \\
\bottomrule
\end{tabular}
}
\caption{Impact of the consistency reward coefficient ($\lambda$) on Llama-3.1-8B-Instruct. We group datasets into LLM-as-a-Judge tasks and MCQ tasks. $\lambda=1.0$ achieves the best balance. The best results are \textbf{bolded}.}
\label{tab:lambda_analysis}
\end{table*}

\subsection{Main Results}
\label{sec:main_results}

As shown in Table \ref{tab:main_results}, across three different LLMs, PA-GRPO yields consistent gains over the strongest baselines, covering both inference-time calibration and training-time debiasing methods. The improvements are particularly concentrated on permutation-sensitive robustness metrics, including Consistency and Consistent Accuracy, while keeping standard Accuracy competitive.

For \textbf{Llama-3.1-8B-Instruct}, PA-GRPO raises MT-Bench accuracy to 77.6\% and consistency to 88.0\%. This representative case suggests that permutation-aware optimization not only improves the headline score but also mitigates decision instability under reordering. This aligns with our motivation that selection bias often manifests as unstable preferences across equivalent permutations.

For \textbf{Qwen3-8B}, the gain becomes more salient in LLM-as-a-Judge evaluations; for instance, JudgeBench accuracy reaches 60.1\%, indicating that PA-GRPO regularizes the model more effectively in settings susceptible to selection bias, preventing it from relying on arbitrary presentation factors (\textit{e.g.}, position or label) for pairwise judgments. Meanwhile, the most notable improvement on MCQ tasks is observed on GPQA, where Consistent Accuracy improves from 46.6\% to 56.7\%, implying that the permutation-group consensus converges more reliably to the ground truth after filtering out permutation noise in complex scientific reasoning. 

For the larger LLM, \textbf{Qwen3-32B}, although the absolute performance is already high, the gains remain persistent; MT-Bench consistency increases to 91.6\%, suggesting that PA-GRPO continues to refine stability even with limited headroom. While a few benchmark-specific metrics show slight regressions, these appear isolated and do not alter the overall trend that enforcing permutation-aware optimization enhances stability and reliability across permutations in both LLM-as-a-Judge and multiple-choice evaluations. A qualitative case study illustrating how PA-GRPO tracks semantic content across permutations, in contrast to vanilla GRPO's surface-level preference, is provided in Appendix~\ref{app:case_study}.

\subsection{Ablation Study}
\label{sec:ablation}
Table \ref{tab:ablation} examines the contributions of the Consistency-Aware Reward $r_{\text{con}}$ and the Cross-Permutation Advantage $A_{\text{PA}}$ on Llama-3.1-8B-Instruct. Adding \textbf{$r_{\text{con}}$} to GRPO explicitly encourages within-group agreement, effectively raising MT-Bench consistency from 80.6\% to 83.1\% without sacrificing correctness. In contrast, enabling \textbf{$A_{\text{PA}}$} reduces permutation-induced reward variance during advantage estimation, which translates into broader gains, such as improving JudgeBench accuracy from 48.2\% to 49.3\%.

Crucially, the two components are complementary: $r_{\text{con}}$ shapes the model toward agreement, while $A_{\text{PA}}$ stabilizes the group-level optimization signal. Their combination yields the strongest overall robustness, boosting PreferenceBench consistent accuracy from 76.3\% to 79.8\% and reaching 75.0\% on TinyMMLU. While minor benchmark-specific accuracy trade-offs are observed, the results confirm that both components are essential for reliable cross-permutation behavior.

\begin{figure}[t]
    \centering
    \includegraphics[trim=8.5cm 15.5cm 24.5cm 7cm, clip,width=\linewidth]{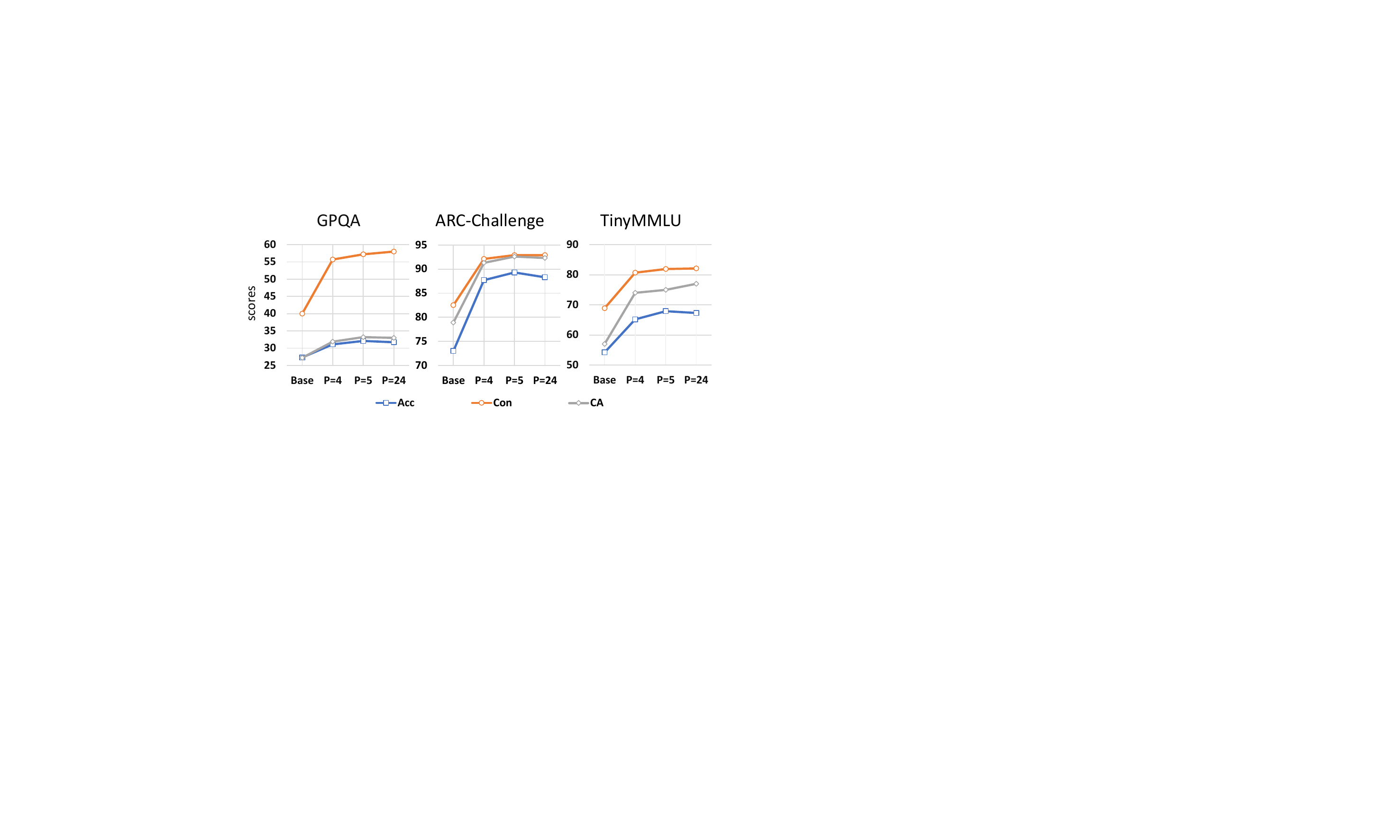} 
    \caption{Impact of Permutation Group Size ($P$). } 
    \label{fig:group_size}
\end{figure}

\subsection{Hyperparameter Analysis}
\label{sec:hyperparams}

We analyze two key hyperparameters of PA-GRPO on Llama-3.1-8B-Instruct: the consistency reward coefficient $\lambda$ and the permutation group size $P$.
Table~\ref{tab:lambda_analysis} summarizes the effect of $\lambda$ across LLM-as-a-Judge and MCQ benchmarks, and Figure~\ref{fig:group_size} reports results for different $P$ on MCQ tasks.

\noindent\textbf{Consistency reward coefficient $\lambda$.}
We evaluate $\lambda$ values of 0.5, 1.0, and 2.0 on Llama-3.1-8B-Instruct. We select $\lambda=1.0$ as it optimally balances performance and consistency, achieving 79.8\% consistent accuracy on PreferenceBench. In contrast, $\lambda=0.5$ sacrifices stability for marginal accuracy gains on MT-Bench, while $\lambda=2.0$ provides no significant performance benefit to justify the stronger regularization.

\noindent\textbf{Permutation group size $P$.}
As illustrated in Figure~\ref{fig:group_size}, for MCQ tasks, we compare $P$ values of 4, 5, and 24. We find that $P=5$, which adds a reverse order to cyclic shifts, outperforms $P=4$ by effectively disrupting adjacency bias. Although expanding to the full set where $P=24$ increases TinyMMLU consistent accuracy from 75.0\% to 77.0\%, the marginal gain does not justify the nearly five-fold computational cost. We therefore adopt $P=5$ for efficiency.

\begin{table*}[ht]
\centering
\resizebox{\textwidth}{!}{
\setlength{\tabcolsep}{3.5pt}
\renewcommand{\arraystretch}{1.25}
\begin{tabular}{l|ccc|ccc|ccc|ccc|ccc|ccc|ccc}
\toprule
& \multicolumn{12}{c}{\textbf{LLM-as-a-Judge}} & \multicolumn{9}{c}{\textbf{Multiple Choice Questions}} \\
\cmidrule(lr){2-13} \cmidrule(lr){14-22}
\multirow{2}{*}{\textbf{Method}} & \multicolumn{3}{c|}{\textbf{MT-Bench}} & \multicolumn{3}{c|}{\textbf{JudgeBench}} & \multicolumn{3}{c|}{\textbf{PreferenceBench}} & \multicolumn{3}{c|}{\textbf{RewardBench}} & \multicolumn{3}{c|}{\textbf{GPQA}} & \multicolumn{3}{c|}{\textbf{ARC-Challenge}} & \multicolumn{3}{c}{\textbf{TinyMMLU}} \\
& Acc & Con & CA & Acc & Con & CA & Acc & Con & CA & Acc & Con & CA & Acc & Con & CA & Acc & Con & CA & Acc & Con & CA \\
\midrule
Base (Direct) & 59.6 & 25.2 & 22.2 & 35.0 & 34.8 & 6.1 & 60.8 & 22.6 & 22.1 & 60.5 & 31.5 & 26.2 & 27.3 & 40.0 & 27.2 & 73.0 & 82.5 & 78.9 & 54.2 & 68.9 & 57.0 \\
Base (CoT)    & 68.7 & 68.2 & 58.0 & 39.0 & 48.5 & 17.1 & 70.3 & 61.3 & 58.1 & 67.5 & 76.2 & 54.8 & 24.5 & 53.9 & 26.6 & 84.6 & 84.2 & 81.1 & 62.4 & 74.3 & 64.0 \\
PA-GRPO (Direct) & 75.1 & 85.5 & 69.3 & \textbf{58.0} & 54.5 & 31.8 & 84.8 & 82.3 & 75.9 & 69.5 & 80.1 & 59.8 & 31.0 & 54.1 & 30.8 & 86.5 & 89.9 & 88.7 & 66.5 & 77.9 & 72.0 \\
\rowcolor{blue!10}
PA-GRPO (CoT) & \textbf{77.6} & \textbf{88.0} & \textbf{71.7} & 57.1 & \textbf{58.3} & \textbf{32.4} & \textbf{86.2} & \textbf{87.2} & \textbf{79.8} & \textbf{71.0} & \textbf{82.7} & \textbf{62.3} & \textbf{32.1} & \textbf{57.2} & \textbf{33.2} & \textbf{89.3} & \textbf{92.9} & \textbf{92.6} & \textbf{67.9} & \textbf{81.9} & \textbf{75.0} \\
\bottomrule
\end{tabular}
}
\caption{Performance comparison of Direct vs. Chain-of-Thought (CoT) inference strategies. We report Accuracy, Consistency, and Consistent Accuracy for both the Base model and PA-GRPO across seven benchmarks.}
\label{tab:cot_analysis}
\end{table*}

\begin{table*}[t]
\centering
\resizebox{\textwidth}{!}{
\setlength{\tabcolsep}{3.5pt}      
\renewcommand{\arraystretch}{1.25} 
\begin{tabular}{l|ccc|ccc|ccc|ccc|ccc|ccc|ccc}
\toprule
& \multicolumn{12}{c}{\textbf{LLM-as-a-Judge}} & \multicolumn{9}{c}{\textbf{Multiple Choice Questions}} \\
\cmidrule(lr){2-13} \cmidrule(lr){14-22}
\multirow{2}{*}{\textbf{Bias Setting}} & \multicolumn{3}{c|}{\textbf{MT-Bench}} & \multicolumn{3}{c|}{\textbf{JudgeBench}} & \multicolumn{3}{c|}{\textbf{PreferenceBench}} & \multicolumn{3}{c|}{\textbf{RewardBench}} & \multicolumn{3}{c|}{\textbf{GPQA}} & \multicolumn{3}{c|}{\textbf{ARC-Challenge}} & \multicolumn{3}{c}{\textbf{TinyMMLU}} \\
& Acc & Con & CA & Acc & Con & CA & Acc & Con & CA & Acc & Con & CA & Acc & Con & CA & Acc & Con & CA & Acc & Con & CA \\
\midrule
Label Bias (Label-only) & 76.7 & \textbf{92.2} & \textbf{72.8} & 56.3 & \textbf{79.0} & \textbf{45.8} & 74.9 & 62.8 & 56.2 & 70.7 & \textbf{87.5} & \textbf{64.4} & 22.0 & \textbf{67.2} & 20.8 & 80.7 & \textbf{95.6} & 81.3 & 63.5 & \textbf{83.8} & 66.0 \\
Position Bias (Order-only)& 77.4 & 85.1 & 69.9 & 55.0 & 45.5 & 27.7 & 74.2 & 66.9 & 57.7 & \textbf{71.6} & 82.0 & 62.6 & 27.9 & 65.0 & 27.7 & 80.9 & 91.9 & 82.8 & 64.3 & 80.5 & 72.0 \\
\rowcolor{blue!10}
Selection Bias (Standard)& \textbf{77.6} & 88.0 & 71.7 & \textbf{57.1} & 58.3 & 32.4 & \textbf{86.2} & \textbf{87.2} & \textbf{79.8} & 71.0 & 82.7 & 62.3 & \textbf{32.1} & 57.2 & \textbf{33.2} & \textbf{89.3} & 92.9 & \textbf{92.6} & \textbf{67.9} & 81.9 & \textbf{75.0} \\
\bottomrule
\end{tabular}
}
\caption{Bias decomposition results under three controlled protocols: Label Bias, Position Bias, and Selection Bias. We report Accuracy, Consistency, and Consistent Accuracy for PA-GRPO across seven benchmarks.}

\label{tab:bias_decomposition}
\end{table*}
\subsection{Chain-of-Thought Analysis}
\label{sec:cot_main}

We further investigate whether the robustness gains of PA-GRPO depend on explicit chain-of-thought (CoT) generation. Table~\ref{tab:cot_analysis} compares four settings on Llama-3.1-8B-Instruct: Base (Direct), Base (CoT), PA-GRPO (Direct), and PA-GRPO (CoT).

The results reveal two complementary effects. First, PA-GRPO improves robustness even under direct decoding, indicating that permutation-aware alignment is internalized into the policy itself. For example, on MT-Bench, consistent accuracy improves from 58.0\% for Base (CoT) to 69.3\% for PA-GRPO (Direct).

Second, explicit reasoning further amplifies these gains. PA-GRPO (CoT) achieves the strongest overall results across most benchmarks, including GPQA, where consistent accuracy rises from 30.8\% under direct decoding to 33.2\% with CoT. This pattern suggests that CoT and permutation-aware training play different but complementary roles: PA-GRPO improves the underlying policy's invariance to label and position changes, while CoT provides additional deliberation that helps the model resolve difficult cases more consistently.

Overall, these results indicate that PA-GRPO does not depend on CoT to achieve robustness, but can benefit further from CoT at inference time when stronger reasoning is available.

\subsection{Bias Decomposition Analysis}
\label{sec:bias_analysis_main}

To better understand residual selection bias after PA-GRPO, we decompose it into three factors: label bias, position bias, and their coupled effect. We evaluate three protocols: \textbf{Label Bias}, which rotates option identifiers while preserving the content-position mapping; \textbf{Position Bias}, which permutes display order while preserving the identifier-content mapping; and \textbf{Selection Bias}, which applies the coupled identifier-position permutation used in our standard evaluation.

As shown in Table~\ref{tab:bias_decomposition}, residual sensitivity to position is often stronger than sensitivity to labels alone, especially on judge-style benchmarks. This trend is clearest on JudgeBench, where consistency is 79.0\% under label-only perturbations but drops to 45.5\% under order-only perturbations. In contrast, MCQ benchmarks generally exhibit weaker label effects; for example, ARC-Challenge reaches 95.6\% consistency under label-only perturbations. These results suggest that residual bias is dataset-dependent and cannot be explained by a single factor.

We further find that the coupled protocol is most aligned with the training objective and is often competitive, especially on PreferenceBench. On PreferenceBench, the standard coupled protocol yields 87.2\% consistency, compared with 62.8\% under label-only perturbations and 66.9\% under order-only perturbations. This is consistent with the design of PA-GRPO, which optimizes consistency under coupled permutations, although the strength of this advantage varies across datasets. The factor-isolated protocols remain informative because they reveal residual sensitivities that are not explicitly disentangled during training. Additional metrics (RStd, CKLD) and a prompt sensitivity analysis are reported in Appendix~\ref{app:bias_diagnostics}.

\section{Conclusion}
\label{sec:conclusion}

In this paper, we proposed PA-GRPO, a novel RL framework designed to mitigate selection bias in LLMs. By integrating a cross-permutation advantage and a consistency-aware reward, our PA-GRPO explicitly enforces semantic invariance across different option arrangements. Extensive experiments on seven benchmarks demonstrate that PA-GRPO significantly outperforms strong baselines, effectively reducing position and label biases while maintaining high reasoning accuracy. This work highlights the efficacy of internalizing consistency constraints to achieve robust and reliable discrete-choice reasoning.

\section*{Limitations}
\label{sec:limitations}

While PA-GRPO proves effective, we acknowledge specific limitations in our current study scope.
First, our method is primarily tailored for discrete-choice tasks (\textit{e.g.}, MCQ and pairwise judgments) where permutation groups are naturally defined.
Extending this consistency-based objective to open-ended generation remains a challenge, as semantic equivalence is harder to quantify in long-form text.
Second, our evaluation is currently centered on English benchmarks and open-weights models.
The interaction between permutation invariance and other systematic artifacts, such as verbosity bias~\cite{dubois2024length} or culture-specific biases in multilingual settings, has not yet been fully explored.
We view our work as a foundational step toward permutation-robust alignment, leaving these broader generalization tests for future research.

\section*{Ethics Considerations}
\label{sec:ethics}

The research presented in this paper is founded on open-source LLMs and utilizes publicly available datasets. Consequently, we do not anticipate that our study will have any direct adverse effects. However, it is crucial to recognize that any generative AI technology, including the contributions of our research, must be implemented with caution to avert potentially harmful outcomes.

\section*{Acknowledgments}
\label{sec:ack}

This work is supported by the National Natural Science Foundation of China (72204087), the Chenguang Program of Shanghai Education Development Foundation and Shanghai Municipal Education Commission (23CGA28), the Shanghai Pujiang Program (23PJC030), Young Elite Scientists Sponsorship Program by CAST (YESS20240562). We also appreciate the constructive comments from the anonymous reviewers.

\bibliography{custom}
\clearpage
\appendix

\section{Statement on the Use of Large Language Models}
\label{sec:ai_statement}

In this study, Large Language Models (LLMs) were employed solely for the purpose of grammatical correction and text refinement to improve readability. They were not involved in the conceptualization of the research, generation of content, or any substantive scientific contribution. The authors retain full responsibility for the content of this manuscript, ensuring its originality and accuracy.

\section{Training Procedure}
\label{app:algorithm}

\begin{algorithm}[h]
\caption{Permutation-Aware GRPO (PA-GRPO)}
\label{alg:pa_grpo}
\begin{algorithmic}[1]
\REQUIRE Policy $\pi_\theta$, old policy $\pi_{\mathrm{old}}$, reference policy $\pi_{\mathrm{ref}}$, permutation set $\Pi$, parameters $P, N, \lambda, \beta$
\FOR{each batch $\mathcal{B}$}
    \STATE Initialize $\mathcal{L}_{\mathrm{batch}} \leftarrow 0$
    \FOR{each instance $x \in \mathcal{B}$}
        \STATE Construct $\mathcal{G}(x)=\{p^{(t)}=\tau_t(x)\}_{t=1}^{P}$
        \STATE Sample $N$ outputs $y^{(t,i)}$ for each $p^{(t)}$ and map them to semantic outputs $z^{(t,i)}$
        \STATE Compute preprocessing rewards $r_{\mathrm{pre}}^{(t,i)}$
        \IF{\textsc{Judge}}
            \STATE Compute consistency rewards $r_{\mathrm{con}}^{(t,i)}$ by pairwise agreement
        \ELSIF{\textsc{MCQ}}
            \STATE Compute consistency rewards $r_{\mathrm{con}}^{(t,i)}$ by majority agreement
        \ENDIF
        \STATE Compute total rewards $r^{(t,i)} = r_{\mathrm{pre}}^{(t,i)} + \lambda r_{\mathrm{con}}^{(t,i)}$
        \STATE Compute advantages $A_{\mathrm{PA}}^{(t,i)}$ using Eq.~\ref{eq:adv}
        \STATE Compute instance objective $\mathcal{L}_x$ using Eq.~\ref{eq:full_obj}
        \STATE $\mathcal{L}_{\mathrm{batch}} \leftarrow \mathcal{L}_{\mathrm{batch}} + \mathcal{L}_x$
    \ENDFOR
    \STATE Update $\theta$ using $\mathcal{L}_{\mathrm{batch}}$
    \STATE Set $\pi_{\mathrm{old}} \leftarrow \pi_\theta$
\ENDFOR
\end{algorithmic}
\end{algorithm}

\section{Benchmark Descriptions}
\label{app:benchmarks}

This appendix briefly describes the seven evaluation benchmarks used in Section~\ref{sec:datasets}.
We group them into (i) pairwise judging benchmarks, where each instance contains an instruction (or question) and two candidate responses with a reference preference label, and (ii) MCQ benchmarks, where each instance is a 4-option multiple-choice question with a provided answer key.
Unless otherwise noted, our permutation-based evaluation protocol (Appendix~\ref{app:data_construction}) is applied to the canonical input format of each benchmark.

\paragraph{MT-Bench.}
MT-Bench~\cite{zheng2023judging} is a controlled, multi-turn instruction-following benchmark designed to evaluate conversational assistants.
It consists of 80 manually written, high-quality multi-turn prompts spanning eight common user categories (\textit{e.g.}, writing, roleplay, extraction, reasoning, math, coding, and two knowledge-oriented categories), with 10 prompts per category.
Each MT-Bench prompt contains two turns, enabling evaluation of context tracking and multi-step dialogue behaviors.
For pairwise judging settings, we use the standard pairwise formulation of MT-Bench prompts where a judge compares two model responses for the same prompt.

\paragraph{JudgeBench.}
JudgeBench~\cite{tan2024judgebench} is a benchmark for evaluating the reliability of LLM-based judges under realistic pairwise comparison settings.
Each example is a question paired with two candidate responses, accompanied by an objective preference label indicating which response is more correct.
The released benchmark includes response pairs generated by strong frontier models, and is designed to probe common judging failure modes such as order sensitivity and bias.

\paragraph{PreferenceBench.}
PreferenceBench~\cite{kim2024prometheus} is an in-domain pairwise ranking benchmark introduced alongside Prometheus~2.
It is constructed by pairing two responses from a pool of multiple candidate responses per prompt, producing a set of 200 prompts with 2,000 labeled response pairs.
In addition, PreferenceBench includes prompt-specific evaluation criteria intended to make the comparison more fine-grained and to stress judges' adherence to explicit rubrics.

\paragraph{RewardBench.}
RewardBench~\cite{lambert2025rewardbench} evaluates reward models (and judge-like scorers) on preference-style comparisons.
Each instance is formatted as a prompt with a human-verified chosen completion and a rejected completion.
The benchmark aggregates structured comparisons spanning several high-level categories, including chat instruction-following, reasoning-focused comparisons, and safety/refusal behavior, aiming to cover both in-distribution and challenging out-of-distribution preference cases.

\paragraph{ARC-Challenge.}
ARC (AI2 Reasoning Challenge)~\cite{clark2018think} is a multiple-choice science QA benchmark, where ARC-Challenge represents the more difficult subset.
Questions are grade-school level science problems that typically require non-trivial reasoning beyond superficial retrieval.
We use the standard ARC-Challenge configuration with its widely used train/validation/test splits and the provided answer key for evaluation.

\paragraph{TinyMMLU.}
TinyMMLU~\cite{polo2024tinybenchmarks} is a compact evaluation set derived from MMLU, intended to approximate MMLU-style performance with far fewer questions.
It consists of a small subsample of MMLU-format questions spanning diverse subject areas, each presented as a 4-option multiple-choice problem.
This benchmark enables faster iteration while retaining broad topical coverage.

\paragraph{GPQA.}
GPQA~\cite{rein2024gpqa} is a graduate-level, domain-expert-written multiple-choice QA benchmark designed to be difficult for non-experts even with access to web search.
All questions are 4-option MCQ drawn from three high-level domains (biology, physics, and chemistry), with curated subsets that vary in strictness and difficulty (including a higher-quality, harder subset).
We follow the standard GPQA evaluation protocol using the released answer key.

\section{Data Construction and Evaluation Details}
\label{app:data_construction}

\paragraph{Training Data Filtering.}
To ensure training efficiency and explicitly target selection bias, we performed pre-processing filtering on the source datasets (Chatbot Arena and MMLU). 
We utilized Llama-3.1-8B-Instruct to infer each training instance under both original and reversed orders (or cyclically shifted orders for MCQ). 
We specifically retained instances where the model yielded inconsistent predictions across permutations, as these samples provide the strongest signal for learning permutation invariance.

\paragraph{Full Permutation Protocol.}
During inference, we strictly evaluated the complete permutation space to calculate consistency metrics:
\begin{itemize}
    \item \textbf{LLM-as-a-Judge Tasks:} For each pairwise comparison, we evaluated all $2!=2$ possible orders: standard (AB) and reversed (BA).
    \item \textbf{MCQ Tasks:} For each 4-option question, we evaluated all $4!=24$ possible arrangements of the options.
\end{itemize}
All reported metrics (Accuracy, Consistency, Consistent Accuracy) are aggregated results based on these fully expanded permutation groups.

\section{Baseline Details and Implementation}
\label{app:baselines}

This appendix provides (i) the core idea of each baseline and (ii) how we implemented it in our experimental pipeline.

\paragraph{PriDe (Debiasing with Prior Estimation).}
PriDe is a label-free, inference-time debiasing method originally proposed for multiple-choice selection robustness~\cite{zheng2023large}.
Let $q$ be the question, option-ID tokens be $\{d_i\}_{i=1}^n$ (\textit{e.g.}, A/B/C/D), and option contents be $\{o_i\}_{i=1}^n$.
PriDe assumes the observed prediction distribution over option IDs can be decomposed into a prior term over $d_i$ and a debiased term over the paired option content (under a permutation)~\cite{zheng2023large}:
\begin{equation}
P_{\text{obs}}(d_i \mid q, x^{I}) \propto P_{\text{prior}}(d_i \mid q)\, P_{\text{deb}}(o_{f_I(i)} \mid q, x),
\end{equation}
where $I$ indexes an option permutation (we use cyclic permutations as in the paper), and $f_I(i)$ maps the $i$-th ID token to the paired option content under $I$.
With cyclic permutations $\mathcal{I}$, PriDe estimates a sample-specific prior by~\cite{zheng2023large}:
\begin{equation}
\begin{split}
P_{\text{prior}}(d_i \mid q) = \text{softmax}\biggl( &\frac{1}{|\mathcal{I}|}\sum_{I \in \mathcal{I}} \\
& \log P_{\text{obs}}(d_i \mid q, x^{I}) \biggr).
\end{split}
\end{equation}
It then averages priors computed on a small estimation subset to obtain a global prior $\tilde{P}_{\text{prior}}(d_i)$, and debiases remaining samples by~\cite{zheng2023large}:
\begin{equation}
P_{\text{deb}}(o_i \mid q, x)\ \propto\ \frac{P_{\text{obs}}(d_i \mid q, x)}{\tilde{P}_{\text{prior}}(d_i)}.
\end{equation}

For each dataset, we select a small subset of evaluation instances as the prior-estimation set.
For each instance, we compute $P_{\text{obs}}(d_i \mid q, x^{I})$ under cyclic permutations (for pairwise judging, $n{=}2$ reduces to swapping).
We estimate per-instance priors using the equation above, average them into a global prior $\tilde{P}_{\text{prior}}$, and then apply the debiasing ratio on the remaining instances using only the default prompt once (followed by renormalization and argmax).

\paragraph{CalibraEval (Label-free calibration for LLM-as-a-Judge).}
CalibraEval treats selection bias as a calibration problem and learns a mapping function $g(\cdot)$ that maps observed probabilities to an unbiased distribution~\cite{li2025calibraeval}:
\begin{equation}
P_{\text{deb}}(t_i \mid I, X_0) = g\!\left(P_{\text{obs}}(t_i \mid I, X_0)\right),
\end{equation}
where $t_i$ is an option-ID token (\textit{e.g.}, A/B) and $X_0$ is the default pairing of tokens and contents.
It optimizes $g(\cdot)$ so that the calibrated judgments remain consistent under swapping positions and/or swapping ID tokens, and proposes a non-parametric order-preserving algorithm (NOA) to solve the resulting optimization~\cite{li2025calibraeval}.

We construct a small unlabeled estimation set and fit the calibration mapping under the corresponding swap or permutation variants required by the target task format. At test time, we apply the learned mapping to the observed probability distribution and renormalize it to obtain the final calibrated decision.

\paragraph{UniBias (Internal attention/FFN manipulation).}
UniBias identifies biased model components (attention heads and FFN vectors) by analyzing their contributions to label prediction, using three criteria: (i) relatedness to label prediction, (ii) bias/imbalance across labels, and (iii) low variance of the injected label information across samples; thresholds are selected via grid search~\cite{zhou2024unibias}.
After identifying biased components, UniBias mitigates bias by masking (eliminating) their influence during inference~\cite{zhou2024unibias}.

We follow the UniBias procedure to build a small support set, identify biased attention heads and FFN vectors under the paper’s criteria and threshold selection, and apply the corresponding masks during evaluation. This keeps the inference-time complexity essentially identical to the original model once the masks are fixed~\cite{zhou2024unibias}.

\paragraph{PIF (Point-wise Intelligent Feedback).}
PIF is an SFT algorithm designed to improve Multiple-Choice Symbol Binding (MCSB).
It constructs negative instances by randomly combining incorrect option contents with all candidate symbols and introduces a point-wise loss to provide direct feedback on these negatives, thereby reducing selection bias and improving accuracy~\cite{xue2024strengthened}.

We implement PIF under the same symbol-binding training principle in both task formats, constructing task-consistent point-wise negatives and optimizing the corresponding point-wise objective. We use the same backbone model and training budget as other training-side baselines for a fair comparison.

\paragraph{GRPO (Group Relative Policy Optimization).}
GRPO is a PPO-style reinforcement learning algorithm that samples a group of outputs for the same prompt and uses relative (within-group) reward signals to optimize the policy, avoiding an explicit value model and reducing memory cost compared to standard PPO~\cite{shao2024deepseekmath}.

We implement GRPO as our direct RL baseline using the same reward design and training budget as PA-GRPO, but treat each permuted prompt as an independent sample and do not enforce cross-permutation consistency.

\section{Training Hyperparameters and Configurations}
\label{app:hyperparams}

We implemented PA-GRPO based on the VeRL framework. Our experiments utilized three backbone models: \textbf{Llama-3.1-8B-Instruct}, \textbf{Qwen3-8B}, and \textbf{Qwen3-32B}.
These models were evaluated across both MCQ and Judge tasks.
To ensure computational efficiency, we leveraged FlashAttention~\cite{dao2022flashattention} for accelerated attention computation and employed Low-Rank Adaptation (LoRA) on all linear layers with rank $r=32$ and $\alpha=64$. 
For Qwen3-8B and Qwen3-32B, we used the non-thinking mode to ensure a consistent decoding setting.
The models were optimized using AdamW~\cite{loshchilov2017decoupled} with a learning rate of 1e-5 for 2 epochs. 
We set the KL regularization~\cite{jaques2019way} coefficient $\beta=0.001$ and the entropy coefficient to 0.001 to stabilize training.
Rollout generation was accelerated using vLLM~\cite{kwon2023efficient} with a sampling size of $N=8$ per prompt variant.
Regarding batch configurations, we set the global batch size to 40 for MCQ tasks and 32 for Judge tasks to accommodate varying memory constraints.
To ensure statistical reliability, all reported training results are averaged over 3 independent runs using different random seeds. For the evaluation phase, we set the temperature to \(0\) to strictly minimize sampling noise and ensure the reproducibility of the selection bias metrics.

We additionally use a length regularization term and a format reward to regulate response length and structure, which helps stabilize outputs during training. The overall reward is jointly determined by the accuracy reward, the permutation-consistency reward, and these auxiliary rewards.

\section{Qualitative Case Study: Vanilla GRPO vs.\ PA-GRPO}
\label{app:case_study}

To provide a concrete qualitative example, we present a case study from our evaluation that contrasts the behavior of vanilla GRPO and PA-GRPO under an answer-order permutation. The semantic contents of the two candidate answers are swapped between the original prompt and the permuted prompt, while the underlying task remains unchanged.

\paragraph{Task Setup.}
We consider a pairwise judgment example from the LLM-as-a-Judge evaluation. In the original prompt, the model is asked to compare two candidate answers to the following query: \emph{``Thomas is very healthy, but he has to go to the hospital every day. What could be the reasons? Can you explain why the above question is interesting?''} In the permuted prompt, the semantic contents assigned to Answer A and Answer B are swapped. A permutation-consistent judge should therefore follow the stronger semantic content rather than preserving the same surface choice across the two prompt variants due to answer labels or positions.

\begin{promptbox}{Original Prompt}
Given a question and two answers, determine which one better answers the question. Please think step by step.

First, put your detailed reasoning and comparison inside \texttt{<think> </think>} tags.

Then, provide your final decision inside \texttt{<answer> </answer>} tags. The answer should be a single letter (A or B).

\vspace{1mm}
\textbf{Query:}

Thomas is very healthy, but he has to go to the hospital every day. What could be the reasons? Can you explain why the above question is interesting?

\vspace{1mm}
\textbf{Answer A:}

There are several possible reasons why Thomas may have to go to the hospital daily even though he is healthy: \ldots

The question is interesting for several reasons: \ldots

\vspace{1mm}
\textbf{Answer B:}

There could be several reasons for Thomas to visit the hospital daily despite being healthy. Some possible reasons include: \ldots

The question is interesting for several reasons: \ldots
\end{promptbox}

The permuted prompt is identical except that the semantic contents of Answer A and Answer B are swapped. This tests whether the model tracks semantic quality rather than answer labels or display position.

\paragraph{Vanilla GRPO.}
Vanilla GRPO produces the same surface choice under both prompt orders, selecting Answer A in both cases.

\begin{promptbox}{Original Response}
\texttt{<think>} \\
Okay, let's tackle this. The user wants to know which answer (A or B) better explains why the question is interesting. The question is about Thomas being healthy but going to the hospital every day. The task is to compare the two answers and determine which one better addresses why the question is interesting. \\

First, I need to understand what each answer says. Answer A provides a list of possible reasons for Thomas's hospital visits and then explains why the question is interesting by listing six points. Answer B also lists possible reasons and then explains the interest in the question with five points. \\

Looking at the structure, both answers address the question's interest. But the user is asking which answer better explains why the question is interesting. So I need to compare the depth and breadth of their explanations. \\

Answer A's explanation of why the question is interesting includes points about paradox, lack of context, variety of explanations, psychological rewards, empathy, and open-endedness. These points are detailed and cover multiple angles: curiosity, deduction, emotional engagement, and the nature of the question itself. \\

Answer B's points are about challenging assumptions, empathy, diverse perspectives, critical thinking, and being a conversation starter. These are also valid, but maybe less comprehensive. Answer A's points are more varied and cover aspects like the psychological aspects (aha moment) and the paradox, which might make the question more intriguing. \\

Additionally, Answer A's explanation is more detailed, with six points, while Answer B has five. The depth of each point in Answer A seems more thorough. For example, Answer A mentions the paradox, lack of information, variety of explanations, emotional engagement, psychological reward, and open-endedness. Answer B's points are also good but perhaps not as comprehensive. \\

So, Answer A's explanation of why the question is interesting is more comprehensive and covers more angles. Therefore, Answer A better answers the question by providing a more detailed and varied explanation of the question's interest. \\
\texttt{</think>} \\
\texttt{<answer>A</answer>}
\end{promptbox}

\begin{promptbox}{Permuted Response}
\texttt{<think>} \\
Okay, let's tackle this. The user wants me to determine which answer (A or B) better addresses the original question. The question is about why Thomas, who's very healthy, has to go to the hospital every day. Then they ask why the question is interesting. \\

First, I need to compare the two answers. Let's look at Answer A. It lists several reasons like work, caregiving, volunteering, medical research, therapy, preventive care, and being a student. Then it explains why the question is interesting, pointing out assumptions, empathy, diverse perspectives, critical thinking, and conversation starters. \\

Answer B also lists possible reasons but includes more specific scenarios like visiting a sick family member, routine tests, clinical trials, working at the hospital, administrative tasks, blood donation, logistical reasons, and even questioning the accuracy of Thomas's health status. The explanation for why the question is interesting includes the paradox, lack of context, variety of explanations, psychological reward, empathy, and open-endedness. \\

Now, the original question is about the reasons for Thomas's daily hospital visits despite being healthy and why the question is interesting. Both answers cover the reasons, but I need to see which one is more comprehensive or better structured. \\

[... omitted several intermediate comparison steps for brevity ...] \\

Overall, both answers are good, but Answer B might be more detailed in the interest part. However, the user might prefer Answer A's structure. \\

In conclusion, both answers are good, but Answer A is more structured and covers the interest part with detailed points. Answer B has more specific scenarios. \\

I think I need to make a decision here. Based on the structure and the thoroughness, Answer A is better. \\
\texttt{</think>} \\
\texttt{<answer>A</answer>}
\end{promptbox}

This behavior indicates that vanilla GRPO preserves the same surface choice under permutation, rather than consistently following the stronger semantic content after the answer contents are swapped.

\paragraph{PA-GRPO.}
PA-GRPO tracks the stronger semantic content across the two prompt variants, selecting Answer A on the original prompt and Answer B on the permuted prompt.

\begin{promptbox}{Original Response}
\texttt{<think>} \\
Okay, let's tackle this. The user wants me to determine which answer (A or B) better explains why the question is interesting. The query is about Thomas being healthy but going to the hospital daily. The task is to compare the two answers provided. \\

First, I need to understand the original question. It's a bit of a paradox because if someone is very healthy, why would they need to go to the hospital every day? The interesting part is the contradiction and the need to explore possible reasons beyond the obvious medical ones. \\

Now, looking at answer A. It lists eight possible reasons, ranging from caring for others, working at the hospital, participating in clinical trials, administrative tasks, blood donation, logistical reasons, and even the possibility that the health claim is incorrect. Then, it explains why the question is interesting by pointing out the paradox, the lack of context requiring speculation, the variety of explanations, the psychological reward of solving it, and the emotional engagement. The answer A seems thorough, covering a wide range of possibilities and explaining the question's interest in detail. \\

Answer B also lists several reasons: work, caregiving, volunteering, medical research, therapy, preventive care, and being a student. Then, it explains the question's interest by highlighting challenging assumptions, encouraging empathy, diverse perspectives, critical thinking, and being a conversation starter. Answer B is also comprehensive but perhaps a bit more concise in its explanations of why the question is interesting. \\

Now, comparing the two. Answer A's explanation of why the question is interesting is more detailed. It breaks down the reasons into six points, each with specific examples. Answer B's points are fewer and more general. For example, Answer A mentions the ``paradoxical nature'' and the ``open-endedness'' which are key aspects of the question's intrigue. Answer B talks about challenging assumptions and empathy, which are valid but perhaps less comprehensive. \\

Additionally, Answer A's list of possible reasons is more extensive, covering scenarios like administrative tasks and logistical reasons, which Answer B doesn't mention. However, Answer B's points about diverse perspectives and critical thinking are also valid. But the question specifically asks why the question is interesting, and Answer A's explanation is more thorough and covers more angles. \\

Another point: Answer A's explanation of the question's interest includes the psychological reward of solving a mystery, which adds depth. Answer B's points are good but perhaps not as detailed. \\

So, considering all that, Answer A provides a more comprehensive and detailed explanation of why the question is interesting, making it the better answer. \\
\texttt{</think>} \\
\texttt{<answer>A</answer>}
\end{promptbox}
\begin{table*}[t]
\centering
\resizebox{\textwidth}{!}{
\setlength{\tabcolsep}{4pt}
\renewcommand{\arraystretch}{1.2}
\begin{tabular}{l|cc|cc|cc|cc|cc|cc|cc}
\toprule
& \multicolumn{8}{c|}{\textbf{LLM-as-a-Judge}} & \multicolumn{6}{c}{\textbf{Multiple Choice Questions}} \\
\cmidrule(lr){2-9} \cmidrule(lr){10-15}
\multirow{2}{*}{\textbf{Method / Setting}} & \multicolumn{2}{c|}{\textbf{MT-Bench}} & \multicolumn{2}{c|}{\textbf{JudgeBench}} & \multicolumn{2}{c|}{\textbf{PreferenceBench}} & \multicolumn{2}{c|}{\textbf{RewardBench}} & \multicolumn{2}{c|}{\textbf{GPQA}} & \multicolumn{2}{c|}{\textbf{ARC-Challenge}} & \multicolumn{2}{c}{\textbf{TinyMMLU}} \\
& RStd & CKLD & RStd & CKLD & RStd & CKLD & RStd & CKLD & RStd & CKLD & RStd & CKLD & RStd & CKLD \\
\midrule
Base & 0.364 & 0.378 & 0.216 & 0.627 & 0.386 & 0.453 & 0.331 & 0.291 & 0.056 & 1.570 & 0.025 & 0.005 & 0.033 & 0.028 \\
GRPO & 0.036 & 0.010 & 0.060 & 0.156 & 0.057 & 0.009 & 0.024 & 0.006 & 0.042 & 1.331 & 0.019 & 0.001 & 0.026 & 0.030 \\
UniBias & 0.047 & 0.005 & \textbf{0.050} & \textbf{0.019} & 0.066 & 0.008 & 0.184 & 0.380 & 0.044 & 1.129 & 0.015 & 0.003 & 0.027 & 0.019 \\
PriDe & 0.078 & 0.012 & 0.107 & 0.023 & 0.051 & 0.006 & 0.029 & 0.002 & 0.036 & 0.983 & 0.014 & 0.002 & 0.029 & 0.031 \\
CalibraEval & 0.186 & 0.074 & 0.263 & 0.163 & 0.260 & 0.158 & 0.294 & 0.212 & 0.085 & 1.765 & 0.018 & 0.003 & 0.019 & 0.025 \\
PIF & 0.061 & 0.008 & 0.154 & 0.154 & 0.383 & 0.667 & 0.088 & 0.016 & 0.063 & 1.431 & 0.021 & 0.004 & 0.018 & 0.016 \\
\rowcolor{blue!10}
PA-GRPO & \textbf{0.019} & \textbf{0.002} & 0.092 & 0.047 & \textbf{0.048} & \textbf{0.005} & 0.019 & \textbf{0.0007} & \textbf{0.012} & \textbf{0.662} & 0.009 & \textbf{0.0005} & \textbf{0.011} & 0.013 \\
PA-GRPO (Modified Prompt) & 0.025 & 0.003 & 0.086 & 0.051 & 0.055 & \textbf{0.005} & \textbf{0.016} & 0.0011 & 0.022 & 0.731 & \textbf{0.008} & 0.0007 & 0.014 & \textbf{0.011} \\
\bottomrule
\end{tabular}
}
\caption{Bias evaluation results using RStd and CKLD (lower is better) on Llama-3.1-8B-Instruct across all seven benchmarks. All rows use the default evaluation prompt unless otherwise noted. The final row reports PA-GRPO under the rewritten prompt shown below. \textbf{Bold} indicates the best (lowest) result among all compared settings.}
\label{tab:bias_eval_all}
\end{table*}
\begin{promptbox}{Permuted Response}
\texttt{<think>} \\
Okay, let me try to work through this step by step. The user is asking which answer (A or B) better addresses the original question. The original question is about Thomas being very healthy but needing to go to the hospital every day. The user also wants to know why the question is interesting. \\

First, I need to compare the two answers provided. Both answers list several possible reasons for Thomas's daily hospital visits. Let me check each part. \\

Starting with the first part: the answers to the question. Answer A lists 7 reasons, including work, caregiving, volunteering, medical research, therapy, preventive care, and being a medical student. Answer B also lists 7 reasons, which include visiting a sick family member, routine tests, clinical trials, working at the hospital, administrative tasks, blood donation, logistical reasons, and the possibility that Thomas isn't actually healthy. \\

Now, looking at the second part: explaining why the question is interesting. Answer A mentions points like challenging assumptions, empathy, diverse perspectives, critical thinking, and conversation starters. Answer B talks about the paradox, lack of context, variety of explanations, the ``aha'' moment, empathy, and open-endedness. \\

I need to see which answer better explains why the question is interesting. Both answers cover different aspects. Answer A focuses on the broader implications of the question, like challenging assumptions and promoting empathy. Answer B emphasizes the paradox and the process of deduction. \\

Wait, looking at the original input: the user's query is ``Thomas is very healthy, but he has to go to the hospital every day. What could be the reasons? Can you explain why the above question is interesting?'' Then the two answers (A and B) are responses to the entire query. So each answer includes both the list of reasons and the explanation of why the question is interesting. \\

Looking at the first part (reasons): Answer A has 7 points, Answer B has 7 points. Both are comprehensive. However, Answer B includes an additional point about the possibility that Thomas isn't actually healthy, which is a bit different. \\

For the second part (why the question is interesting): Answer A lists 5 points, Answer B lists 6. Answer B's explanation seems more detailed and covers more aspects. \\

So Answer B is more comprehensive in the first part. \\

In the second part, Answer B's explanation is more detailed. \\

In conclusion, Answer B provides a more comprehensive and detailed explanation of both parts of the question, making it the better answer. \\
\texttt{</think>} \\
\texttt{<answer>B</answer>}
\end{promptbox}

This example qualitatively illustrates the behavioral difference between the two methods: vanilla GRPO may preserve a surface preference across permutations, whereas PA-GRPO more reliably follows the semantically stronger answer.
\section{Bias-Specific Evaluation Metrics and Prompt Sensitivity Analysis}
\label{app:bias_diagnostics}

To further examine whether the gains of PA-GRPO reflect reduced selection bias rather than overfitting to a specific prompt template, we report two additional bias-oriented metrics together with a prompt-variation experiment. All experiments in this section are conducted on \textbf{Llama-3.1-8B-Instruct}.

In addition to the Accuracy, Consistency, and Consistent Accuracy metrics used in the main paper, we introduce two complementary metrics (lower is better) that directly quantify the degree of selection bias from a distributional perspective:
(1) \textbf{RStd (Recall Standard Deviation)}~\cite{zheng2023large}, which measures the dispersion of per-choice recalls across answer IDs; a lower RStd indicates less uneven preference over answer labels; and
(2) \textbf{CKLD (Choice Kullback-Leibler Divergence)}~\cite{choi2025mitigating}, which measures the KL divergence between the predicted choice-label distribution and the ground-truth label distribution; a lower CKLD indicates closer distributional alignment.

Table~\ref{tab:bias_eval_all} reports the RStd and CKLD results across all seven benchmarks. The last row additionally reports PA-GRPO under a substantially rewritten evaluation prompt to test prompt sensitivity.

PA-GRPO achieves the lowest RStd and CKLD values on most benchmarks, indicating both reduced dispersion in per-choice recalls and better alignment between the predicted and ground-truth choice-label distributions. UniBias is slightly better on JudgeBench under these two metrics, but this advantage does not translate into better overall consistency in Table~\ref{tab:main_results}. This comparison suggests that explicitly suppressing a local bias mechanism is not always sufficient to preserve broader reasoning quality across datasets.

\begin{promptbox}{Default Prompt}
Given a question and two answers, determine which one better answers the question. Please think step by step.

First, put your detailed reasoning and comparison inside \texttt{<think> </think>} tags.

Then, provide your final decision inside \texttt{<answer> </answer>} tags. The answer should be a single letter (A or B).
\end{promptbox}

\begin{promptbox}{Modified Prompt}
Please review a given question alongside two provided answers, and evaluate which response addresses the prompt more effectively. You must think through your decision step by step.

Evaluation Guidelines

Focus on Quality: Assess the answers based on their accuracy, relevance, completeness, and clarity.

Maintain Strict Objectivity: Do not let external factors influence your judgment. Remain unbiased with respect to the order in which the answers are presented and the labels assigned to them (A or B).

Begin by placing your detailed reasoning and comparative analysis within \texttt{<think> </think>} tags.

Afterward, output your final verdict inside \texttt{<answer> </answer>} tags. The final answer must be exactly one letter: A or B.
\end{promptbox}

Compared with the default-prompt row in Table~\ref{tab:bias_eval_all}, the modified-prompt row remains close across all seven benchmarks. This result suggests that the improvement is not tied to a particular surface template, but instead reflects a more stable reduction in selection bias.

\end{document}